\documentclass[letterpaper]{article}
\usepackage{aaai}
\usepackage{times}
\usepackage{helvet}
\usepackage{courier}
\usepackage{graphicx}
\usepackage{times}
\usepackage{latexsym}
\usepackage{booktabs}
\usepackage{amsmath}
\usepackage{amssymb}
\usepackage{graphicx}
\usepackage{bm}
\usepackage{CJKutf8}
\usepackage{amsmath}
\usepackage{url}
\usepackage{tabularx}
\usepackage{subfig}
\usepackage{multirow}

\usepackage{color}
\usepackage[ruled,linesnumbered]{algorithm2e}
\begin{document}
%
\title{MetaAID: A Flexible Framework for Developing Metaverse Applications via AI Technology and Human Editing}
\author{
Hongyin Zhu\\
Department of Computer Science and Technology, Tsinghua University, Beijing, China\\
zhuhongyin2020@mail.tsinghua.edu.cn\\
}
\maketitle
\begin{CJK*}{UTF8}{gbsn}
\begin{abstract}
Achieving the expansion of domestic demand and the economic internal circulation requires balanced and coordinated support from multiple industries (domains) such as consumption, education, entertainment, engineering infrastructure, etc., which is indispensable for maintaining economic development. Metaverse applications may help with this task and can make many industries more interesting, more efficient, and provide a better user experience. The first challenge is that metaverse application development inevitably requires the support of various artificial intelligence (AI) technologies such as natural language processing (NLP), knowledge graph (KG), computer vision (CV), and machine learning (ML), etc. However, existing metaverse application development lacks a lightweight AI technology framework. This paper proposes a flexible metaverse AI technology framework MetaAID that aims to support language and semantic technologies in the development of digital twins and virtual humans. The second challenge is that the development process of metaverse applications involves both technical development tasks and manual editing work, and often becomes a heavyweight multi-team collaboration project, not to mention the development of metaverse applications in multiple industries. Our framework summarizes common AI technologies and application development templates with common functional modules and interfaces. Based on this framework, we have designed 5 applications for 3 industries around the expansion of domestic demand and economic internal circulation. Experimental results show that our framework can support AI technologies when developing metaverse applications in different industries.
\end{abstract}

\section{Introduction}

Metaverse applications are designed to map real-world entities or sensations into the virtual world, thereby generating parallel virtual avatars. The blueprint of metaverse \cite{sparkes2021metaverse,park2022metaverse,wang2022metasocieties} is to build a digital twin of the real world. Metaverse applications may make various industries more interesting, more efficient, and provide a better user experience, which will help the expansion of domestic demand and the economic internal circulation. Metaverse application needs the support of AI technologies to realize smarter and more interesting functions. Besides, AI technologies may reduce the cost of system development and operation and maintenance (O\&M). AI has a silent but important role in the foundation and development of the metaverse \cite{huynh2022artificial}. However, in terms of technical support, there is a lack of lightweight frameworks for the AI tech stack. Due to the unclear connection between applications in different industries, applications tend to be developed separately, resulting in a lack of collaboration and information interconnection. The expansion of domestic demand and the economic internal circulation is not limited to one industry, but also requires the balanced and coordinated development of multiple industries (domains) such as consumption, entertainment, education, engineering infrastructure, technological innovation, etc. The coordinated development of different industries is inseparable from the interconnection of information. It is well known that there is a lot of duplication of code between applications. 

Most AI frameworks focus on a single task, such as text classification, sequence labeling, information extraction, knowledge graph reasoning, etc. When it comes to developing an application, it simply requires multiple AI frameworks, resources, front-end and back-end development, system operation and maintenance, etc. When multiple industries are involved, different development teams are required to complete the applications separately. This also leads to less collaboration between different industries. Different from 
prior works, we propose MetaAID, a flexible metaverse framework that uses \textbf{AI} technology and human e\textbf{d}iting to support the collaborative development of applications in multiple industries. This framework is lightweight and flexible and can provide rich language and semantic technical support for developing digital twins and virtual humans. This framework can also be seen as the integration, consolidation, reorganization, and extension of existing frameworks. A digital twin \cite{rasheed2020digital} is a virtual representation of an object or system that spans its lifecycle and uses simulation, machine learning, and reasoning to help decision-making. A virtual human (or digital human) \cite{magnenat2005virtual} is a simulation of a person on a computer that can be used for interaction between human and machine intelligence. 

There are 3 main challenges in implementing this framework. (1) Accumulation and organization of the tech stack. Summarizing and implementing the AI tech stack required for application development is a lengthy process that may take years to complete. This includes the process of collecting and comparing technology options (solutions) and finally, we adopt the most agile and extensible approach. (2) The repository of human-machine collaboration templates. Technical templates or code snippets that can be widely used by developers are essential, such as templates for application development and templates for different programming languages. (3) Framework evaluation. Evaluating the framework by developing online applications can generate direct results. 
For the first challenge, we collect various technical frameworks and solutions over 8 years. For the second challenge, we organize resources to facilitate high-quality human editing work, including summarizing reusable templates and solutions to common problems. For the third challenge, we have multiple applications in different industries developed by one developer using this framework. 

Digital twin technology can be used to map the real-world entities into the virtual world, such as stores, merchandise, teachers, etc. Virtual reality (VR) and augmented reality (AR) aim at the first step in mapping entities into the virtual world, such as dealing with the use of 3D scanners to create human shapes in 3D graphics. Our framework focuses on the second step, which is to provide the virtual avatar with technical support (intelligence) in terms of language and semantics.

Achieving the expansion of domestic demand and economic internal circulation requires balanced and coordinated support from multiple industries. Applications in different industries are interconnected in technical support and data analysis. As an illustrative example, we apply the MetaAID framework to 3 industries that are indispensable for the expansion of domestic demand and economic internal circulation, i.e., entertainment (NLP-based entertainment), online education (AI subjects and English learning), daily consumption (food and beverages). The metaverse platform requires not only machine programs but also human editing of high-quality and personalized content. The advantage of machine programs lies in their computational efficiency, while the advantage of humans lies in their wealth of creativity and knowledge. Our framework aims to integrate technology with human editing work, emphasizing not only technology but also the win-win and mutual progress of humans and technology. 

To evaluate this framework, we develop 5 applications on different platforms, including iOS apps, online websites, and WeChat mini-programs. The innovations of this paper are as follows.

1. We propose MetaAID, a flexible metaverse framework that uses AI technology and human editing to support the collaborative development of metaverse applications in multiple industries. 

2. We build a repository of human-machine collaboration templates to support agile human-machine content creation. 

3. To evaluate the features of this framework, we develop 5 applications in 3 industries around the topic of the expansion of domestic demand and economic internal circulation.

\section{Related Work}
\subsection{AI Technology}
Artificial intelligence \cite{russell2002artificial} is the simulation of human intelligence processes by machines, especially computer systems. The concept of artificial intelligence consists of multiple disciplines, e.g., knowledge representation and reasoning, commonsense, machine learning, natural language understanding and generation, machine vision, psychology, philosophy, etc. 

Machine learning is the process of using mathematical models of data to help a computer learn without direct instruction. Machine learning algorithms have great potential in metaverse applications. These methods can be divided into supervised learning \cite{bishop:2006:PRML}, semi-supervised learning \cite{van2020survey}, unsupervised learning \cite{bishop:2006:PRML}, self-supervised learning \cite{liu2021self} and reinforcement learning \cite{franccois2018introduction}. Models can also be divided into generative models and discriminative models \cite{DBLP:books/daglib/0087929}. These methods can also be divided into traditional machine learning methods and deep learning \cite{goodfellow2016deep} methods. In recent years, neural network models have been widely studied and applied. With the development of computational power and optimization algorithms, researchers can train deep and large neural network models. These models achieve state-of-the-art (SOTA) performance on perceptual and cognitive tasks.

Due to space limitations, considering the scope of this paper, we mainly introduce some frameworks in natural language processing, knowledge graph, computer vision, and deep learning. NLP frameworks are designed to process textual data efficiently, enabling machines to directly understand the unstructured text. There are several representative toolkits that play different roles in text processing, understanding, generation, etc., for example, transformers\footnote{https://github.com/huggingface/transformers}, Stanford CoreNLP \cite{manning2014stanford}, Spacy\footnote{https://spacy.io/}, NLTK \cite{bird2004nltk}, Gensim \cite{vrehuuvrek2011gensim}. KG frameworks aim to build machine-readable knowledge graphs from heterogeneous data and to construct linked open data (LOD) through the semantic web technology standards. There are several representative toolkits under this tech stack that deal with knowledge graph construction, representation, reasoning, question answering, etc., for example, RDFLib \cite{krech2006rdflib}, Jena \cite{mcbride2002jena}, BLINK \cite{wu2019zero}, OpenKE \cite{DBLP:conf/emnlp/HanCLLLSL18}, etc. 

Computer vision frameworks aim to enable computers and systems to derive high-level understanding from digital images, videos, and other visual inputs. There are several representative toolkits that deal with image classification, object detection, face recognition, semantic segmentation, etc., for example, OpenCV \cite{bradski2000opencv}, YOLO \cite{redmon2016you}, SimpleCV\footnote{http://simplecv.org/}, Caffe \cite{jia2014caffe}, etc. Deep learning frameworks are used to build deep neural networks and support parallel training of large-scale data and can train large models with tens of billions of parameters. Representative frameworks are Tensorflow \cite{abadi2016tensorflow}, Pytorch \cite{paszke2019pytorch}, Keras\footnote{https://keras.io/}, etc. In terms of metaverse frameworks, some companies, such as Meta Platforms (Facebook), Microsoft, and Roblox, have gradually built metaverse platforms, but there are few lightweight metaverse frameworks. Zhu \cite{zhu2022metaonce} proposes the MetaOnce framework for adding more constraints and game theory to entities, relations, and events in metaverses. They introduce multi-scene graphs and propose the entity-relation-event game rule controller. 

\subsection{Metaverse Applications}
The Metaverse system \cite{sparkes2021metaverse} is mainly used in 3 scenarios. In terms of content distribution platforms, applications mainly cover the fields of games, social networks, sports, etc. In terms of display platforms and devices, it mainly includes smart wearable devices, VR, and AR. In terms of software and hardware technical support, it mainly includes blockchain, 5G networks, artificial intelligence (AI), cloud computing, chip design, etc.

There are some well-known metaverse systems in the world. There are different services in Facebook's metaverse platform \cite{sparkes2021metaverse,wilson2012review}, covering social networks, games, work scene, collaboration, productivity, etc. Microsoft's metaverse system \cite{kim2021advertising} includes digital twin-based games, augmented reality glasses, a mixed reality collaboration platform, cloud computing services, etc. Roblox is committed to providing users with the tools and technologies to create virtual worlds \cite{lee2021metaverse}. The content covers educational innovation, games, etc. Epic Games \cite{kaur2021metaverse} is working on building the metaverse into game systems. They build a high-fidelity digital human platform based on Unreal Engine in the cloud. Different from the above platforms, our metaverse framework is designed to provide AI technologies when developing virtual humans and digital twins. 
\section{Approach}
In this section, we first provide an overview of the MetaAID framework and then describe our approach to developing applications in 3 industries.
\subsection{Framework Overview}
\begin{figure}[htbp!]
\centering
\includegraphics[width=2.5in]{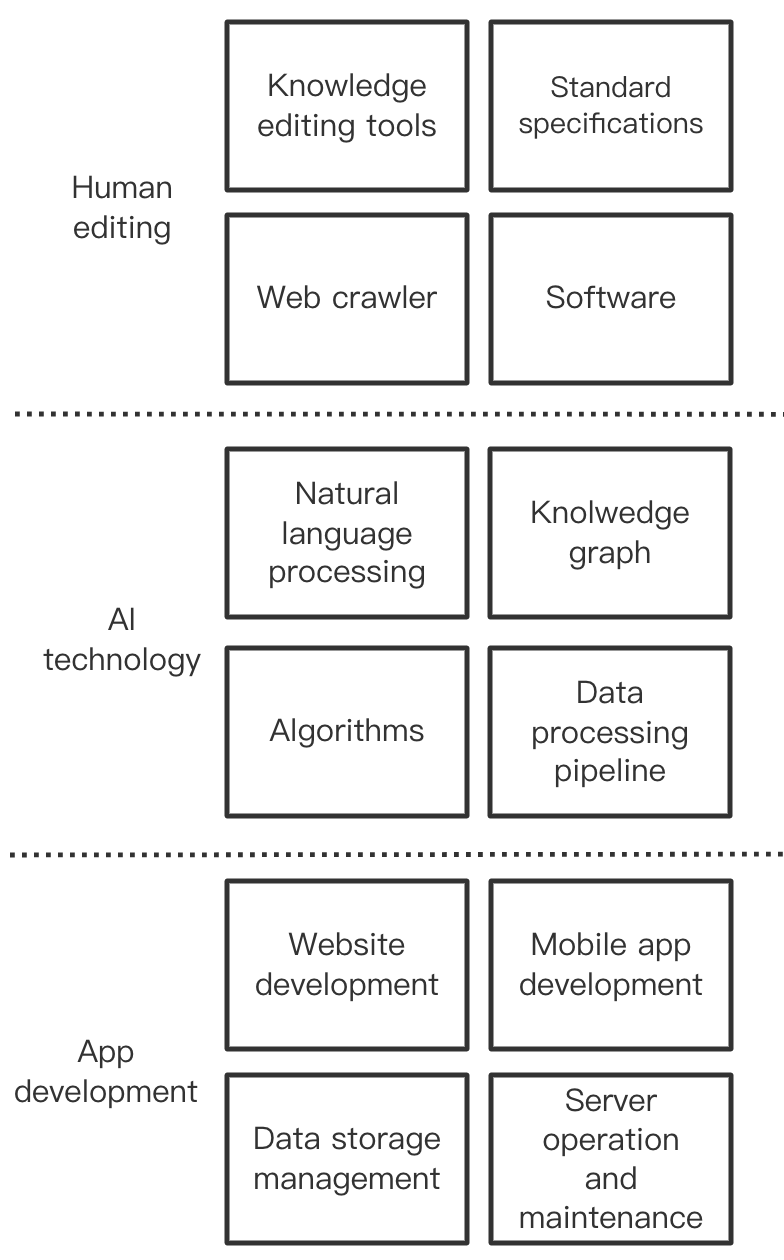}
\caption{Framework overview\label{example1}}
\end{figure}
Our framework is divided into 3 layers, i.e., the human editing layer, AI technology layer, and app development layer, as shown in Figure \ref{example1}. These layers are divided according to technical characteristics, including their tasks, required background knowledge, and roles in the system development process. Efficient collaboration can be achieved between different layers, and modules at different layers are responsible for different tasks. It can also be simply seen as a tech stack similar to some technology companies. A tech stack is a combination of technologies that a company uses to build and run an application or project. Different from a company, our framework is portable, highly open, and customizable, with greater growth potential. 

The bottom layer is the app development layer, which aims to efficiently develop applications, including website development, mobile application development, data storage management, server operation and maintenance. Currently, the website development module only supports the development of monolithic architecture and will be upgraded to support distributed monoliths and microservices applications in the future. This module mainly includes front-end and back-end technical templates, technical usage instructions, bug solving instructions, etc. Mobile application development aims to develop iOS, Android, and WeChat mini-program. Due to the different development languages and operating mechanisms of different platforms, app interface development is different from website front-end development. We summarize some templates for mobile app development. Mobile apps and websites can share the same backend services. Data storage management is designed to manage the data center for private and federated analysis. In our framework, application data of all industries are stored separately in the same data center, which not only ensures the independence and privacy of databases of different industries but also facilitates the collaboration and joint analysis of multi-industry data. The databases we use mainly include relational databases, graph databases, and simple JSON-based databases. The operation and maintenance module includes instructions for using common tools such as Docker, Tomcat, hardware management, network management, and shell scripts.

The middle layer is the AI technology layer, including NLP models, KG technology, algorithms, and data processing pipelines. Our NLP models include text classification, text generation, sequence labeling, information extraction, question answering, machine reading comprehension, dialogue, etc. KG technology module contains ontology construction, entity linking, entity expansion, KG representation, KG reasoning, KG question answering, etc. The algorithm module contains APIs for hundreds of algorithms for solving some complex problems, such as graph search problems, dynamic programming, etc. The data processing pipeline includes not only some data cleaning algorithms but also some traditional machine learning algorithms including supervised and unsupervised models, such as the SVM, K-means, decision tree, random forest, etc. 

The upper layer is the human editing layer, which allows people to edit more personalized and high-quality content. This layer consists of knowledge editing tools, standard specifications, web crawlers, and software. Knowledge editing tools include data annotation tools, domain dictionary construction tools, protege, tools for updating knowledge graphs, etc. The standard specification aims to help editors produce high-quality, positive, and user-satisfying material, and includes principles and presentation techniques, e.g., video tutorial recording tips, PPT presentation tips, HTML editing tips, etc. Web crawlers are designed to fetch customized content from the Internet to improve work efficiency. Software is a necessary part to help with the manual editing process, such as optical character recognition (OCR) software, UI design software, screen recording software, etc. Based on this framework, we build 5 applications in 3 industries. Next, we will describe the problems encountered during the application building process and our solutions. 

\subsection{Entertainment Industry Application}
NLP technologies are widely used in public opinion monitoring, scientific literature mining, news mining,  biomedicine, etc. NLP techniques have achieved good results on tasks such as dialogue, question answering, and machine reading comprehension, and have surpassed human-level performance on some datasets. People look forward to interacting happily with intelligent machines and enjoying the convenience brought by technological progress. 

%
While NLP techniques are widely used, especially for text classification, sequence labeling, and text generation tasks, few mobile applications use NLP techniques to entertain users. We develop a Chinese analysis app ``What you say" (你说啥子), designed to automatically analyze human language and generate responses on mobile phones. This technology can be used to develop virtual humans in metaverse applications. This app contains machine reading comprehension and emotion recognition functions. The machine reading comprehension task is defined as follows. Given a document $t$ and a question $q$, the model answers the question by extracting the text spans in the document, which is also known as extractive question answering. The emotion recognition task is defined as follows. Given a text $t$, the model aims to predict the most likely emotion type.
\begin{align}
\label{emotioneq}
c_i = argmax(p(c_1), p(c_2), ..., p(c_n))
\end{align}
where $p(c_i)$ is the probability of emotion $c_i$.

\begin{figure}[htbp!]
\centering
\includegraphics[width=3.2in]{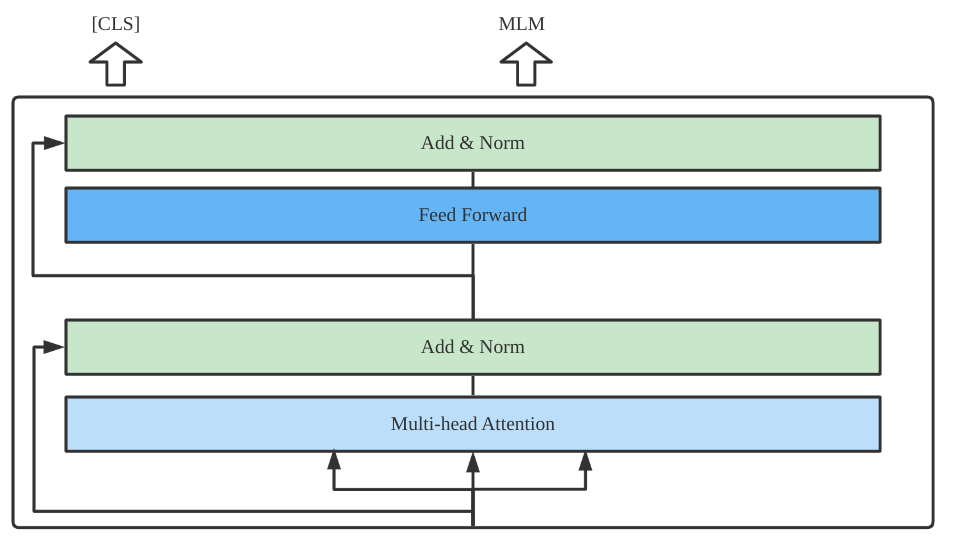}
\caption{Transformer block architecture\label{transformer}}
\end{figure}

The challenge lies in NLP models and mobile app development. For the first challenge, we use the NLP module in the AI technology layer. We adopt BERT \cite{DBLP:conf/naacl/DevlinCLT19}, which is further fine-tuned by \cite{zhao2019uer} to deal with these two tasks. , As shown in Figure \ref{transformer}, BERT is composed of transformer blocks \cite{vaswani2017attention}, and the formula for the transformer block is as follows. 

\begin{align}
\label{11}
h = softmax(\frac{QK^T}{\sqrt{d_k}})V
\end{align}
The above equation shows the scaled dot-product attention where Q,K,V are the query, key, value matrices. The input consists of queries and keys of dimension $d_k$, and values of dimension $d_v$. 
\begin{align}
\label{22}
H_{mul} = (h_1 \oplus h_2 ... \oplus h_n)W^O
\end{align}
where $W^O$ is the parameter matrix.
\begin{align}
\label{33}
h_{norm} = LayerNorm(V + SubLayer_1(V))
\end{align}
The above equation is a residual connection \cite{he2016deep} around the sub-layer, followed by layer normalization. $SubLayer_1(\cdot)$ is the multi-head self attention function. 
\begin{align}
\label{44}
h_{ffn} =max(0,h_{norm} W_1+b1)W_2+b_2
\end{align}
$W_1$ and $W_2$ are liner transformations. $b_2$ is the bias. The linear transformations are the same across different positions, and they use different parameters from layer to layer.
\begin{align}
\label{55}
h_{out} = LayerNorm(h_{norm}+SubLayer_2(h_{norm}))
\end{align}
The above equation is another layer normalization and residual connection. $SubLayer_2(\cdot)$ is the position-wise feed-forward network. 

For the second challenge, we develop an iOS app using the mobile app development module to facilitate user interaction. In terms of cloud services, since the training and inference of deep learning models require GPUs, we built cloud services using server operation and maintenance modules. 

\subsection{Education Industry Application}
With the development of science and technology, the problem of uneven distribution of educational resources is being alleviated. However, the dissemination of scientific and technological knowledge on frontier research topics needs to be further resolved. There needs to be more sharing of learning experiences from novice to expert. 

We build 3 types of educational applications, corresponding to knowledge integration, knowledge dissemination, and knowledge testing. This technique can be used to develop educational virtual humans. For knowledge integration, we develop an online knowledge resource aggregation website ``Top bros online" (小牛人在线). Existing online education websites rarely consider searching other educational resource websites, and return all the results to users after aggregating, because different educational websites are separated. The difficulty of this problem lies in the integration of multi-source knowledge on the Internet. We build a website to aggregate massive information technology videos and construct a knowledge graph to support comprehensive resource SPARQL. The data comes from multiple Chinese online education websites, e.g., chinahadoop.cn, study.163.com, xue.taobao.com, etc.

For knowledge dissemination, we manually record hundreds of video tutorials. The challenge lies in the production of scarce video resources. We record the videos on AI disciplines, such as pattern recognition, natural language processing, algorithm and data structure, matrix theory, probability theory and mathematical statistics, etc. We develop a WeChat mini-program ``Theory bros online" (师兄在线) where users can select a video playlist of a course to watch the entire video list. The app uses manually recorded videos to address the spread of AI theoretical knowledge that is difficult for beginners to understand. In terms of structured knowledge, our knowledge card module extracts structured knowledge points to make the books more interesting and easier to learn. First, we scan paper books and use OCR to recognize images, and then use multi-strategy information extraction to extract structured data. 

For knowledge testing, we develop an online educational website ``English doctor" (英语医师) which aims to test users' knowledge of English. A major feature of this website is that we organize English knowledge from keywords, knowledge points, question types, etc. 
Our website has the function of reciting words and knowledge testing. This site uses questions from common exams such as CET-4, CET-6, and the graduate entrance examination to test users' knowledge. On this English testing system, users only need to answer the questions, and the system will automatically verify the correctness of the answers. The system will record each user's wrong answers and unremembered words, increasing the probability of checking these knowledge points, and allowing users to deepen their memory. These features are supported by knowledge graph technology, and we construct an English test knowledge graph.

\subsection{Consumption Industry Application}
While the e-commerce and food delivery industries are booming, the market share of physical stores declines. However, physical store consumption has its irreplaceable advantages, such as allowing people to better integrate into society, experience different environments, and buy or learn unexpected new products. Due to the lack of attractiveness and interest in offline consumption scenarios, people are more inclined to choose convenient online consumption, especially when they can stay at home and within reach. 

We propose ``Talk over drinks" (谁来煮酒) app, which aims to increase the fun of food and beverages consumption. This app promotes offline consumption, drives the market growth of related industries, and helps to enhance interpersonal relationships and social interaction. This technology can be used to model digital twins for physical stores or products in the metaverse application. By integrating multi-source and multi-modal data of search engines, as well as store review websites such as dianping.com and meituan.com, etc, it facilitates users to make search queries and comprehensive comparisons to enhance their leisure experience. Meanwhile, this app can recommend interesting content. 

\begin{figure}[htbp!]
\centering
\includegraphics[width=3.2in]{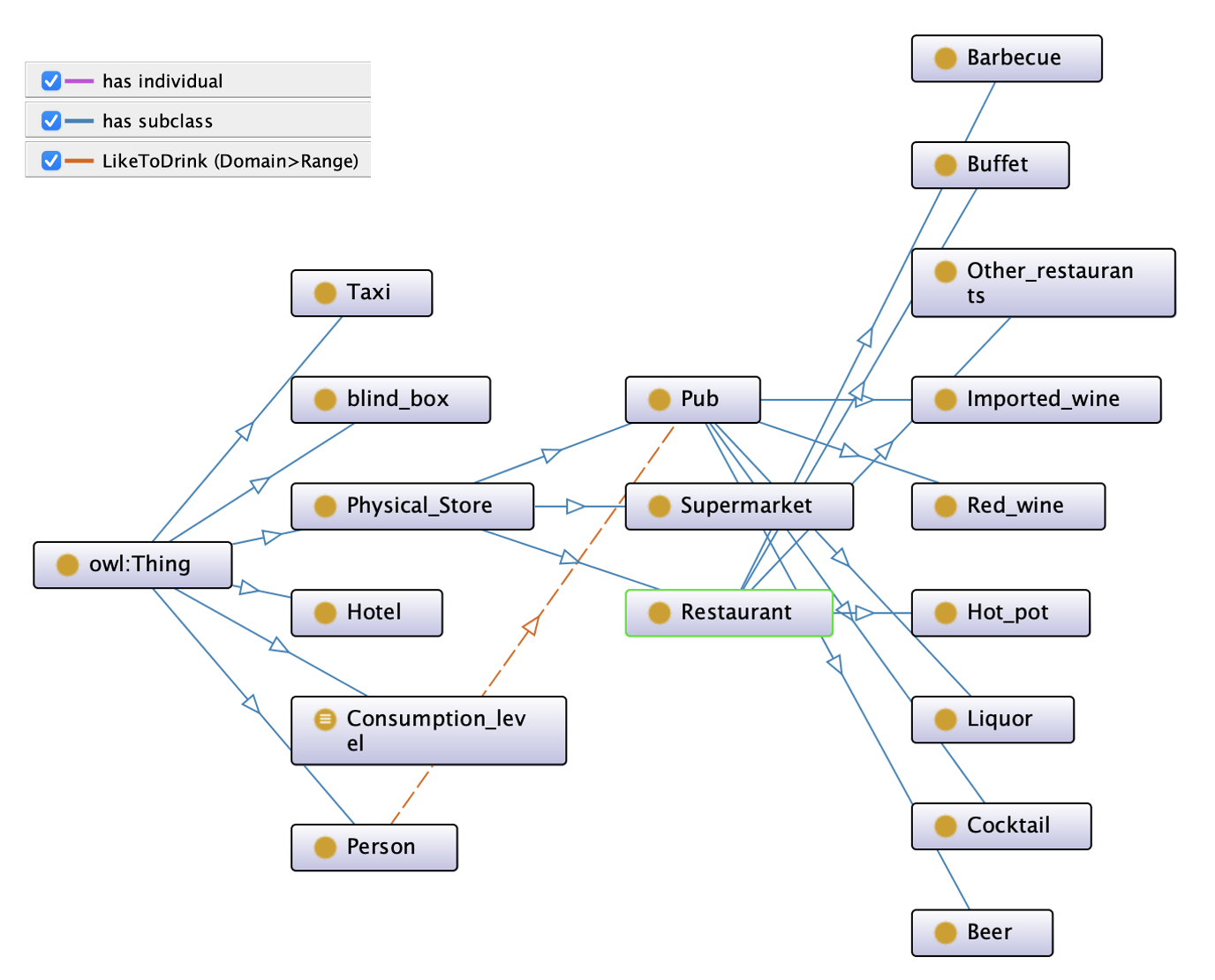}
\caption{Ontology of the MCKG\label{onto}}
\end{figure}

The difficulty of KG construction lies in several aspects, such as data acquisition, ontology construction, and information extraction. We use the KG technology module in the AI technology layer. We first obtain raw data from search engines, then use the information extraction algorithms to extract structured information and build a knowledge graph. 
The content of food and beverages is not monotonous but consists of many aspects. We propose the multi-perspective consumption (MC) KG and divide the application into the following aspects, i.e., classification of alcoholic beverages, celebrities and alcoholic drinks, late-night snack, physical stores, hotels, transportation, and blind boxes. For the ontology, as shown in Figure \ref{onto}, the top-level concepts contain Person, Consumption level, Hotel, Physical store, Blind box, Taxi.

Classification of alcoholic beverages contains a detailed description, such as liquor, imported wine, red wine, champagne, cocktail, beer, etc. For the above alcoholic beverages, we curate content from the Internet or books. Celebrities and alcoholic drinks describe the relation of (celebrity, likeToDrink, alcoholic drinks ). We first created a list of celebrities and a list of common alcoholic beverages, which were then automatically linked to the Baidu encyclopedia through a program. We extract the relations (celebrity, likeToDrink, alcoholic drinks ) from search engines and keep the evidence of these facts. We first find webpages that mention both celebrities and alcoholic drinks, then use an information extraction algorithm to extract the facts where celebrities and alcoholic beverages co-occurrence in the same passage, and select the frequently occurring facts. Based on the output candidate facts, we further perform a manual check to verify the correctness of the facts. 

Physical stores are divided into bars and restaurants. We manually set seed entities for automatic data acquisition and entity linking. 
We used dianping.com to find highly rated stores and then organized them by geographic location. In terms of hotels, we select hotels with high ratings on ctrip.com and organize them by geographic location. Users can tap the phone screen to view detailed information and navigate to other well-known apps. 

\section{Experiment}
\subsection{Setup}
We developed 5 applications in 3 industries based on this framework and invited users to test them. These applications run on 3 platforms, i.e., the website, iOS, and WeChat mini-program. These cloud services run on an AMD Ryzen 5 1500X Quad-Core Processor @ 3.5GHz (Mem: 16G) and 1 Tesla T4 GPU (16G). 
\subsection{Results of the Entertainment Industry Application}
\begin{figure}[h]
\centering
\includegraphics[scale=0.3]{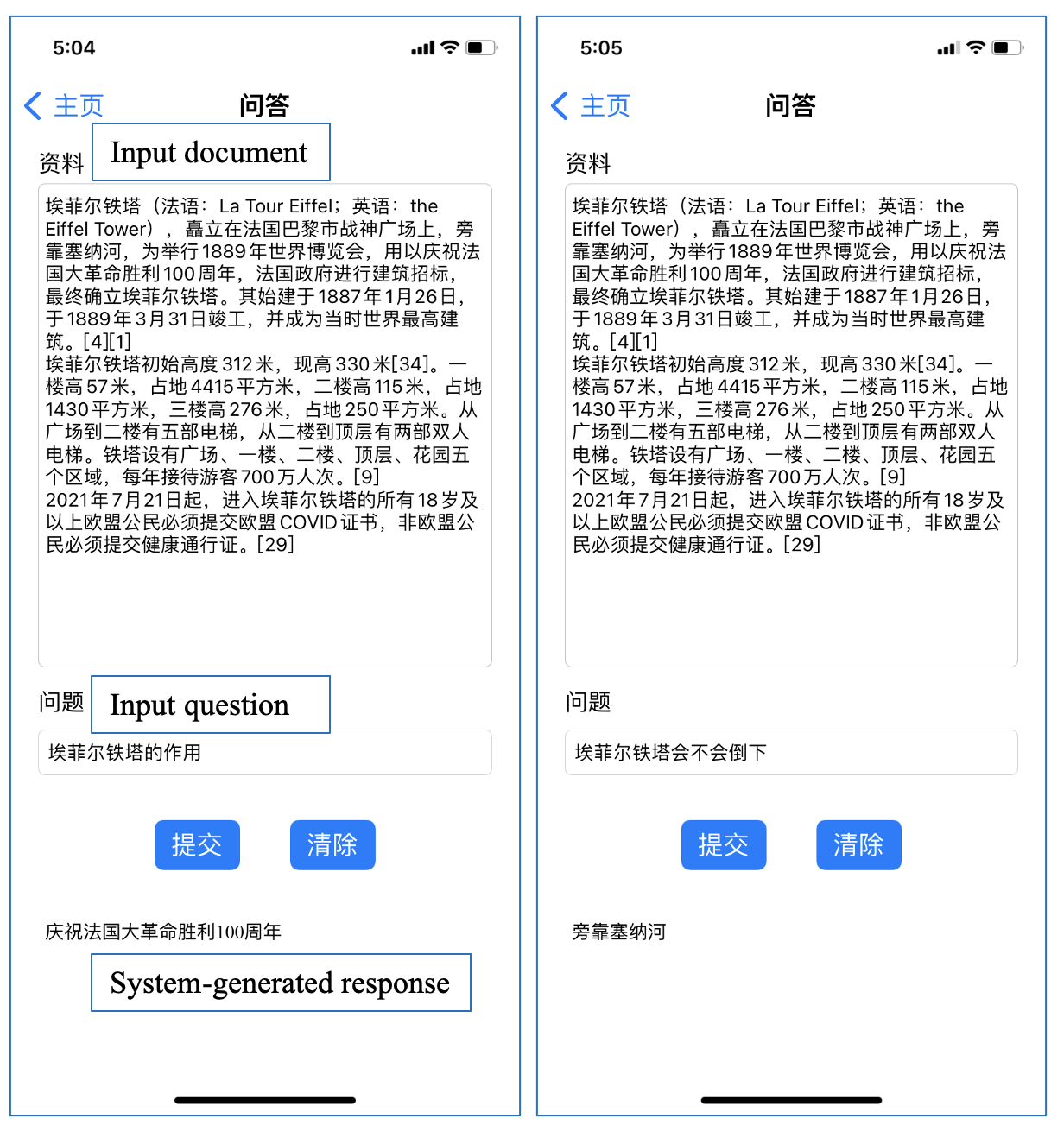}
\caption{Examples of MRC function}
\label{fig:mrc}
\end{figure}

Figure \ref{fig:mrc} shows the results from the machine reading comprehension (MRC) function and the emotion recognition function. For the MRC function, the user inputs a document about ``the Eiffel Tower'', then asks different questions, and the system returns corresponding answers based on the document. For example, the input question is ``what is the meaning of the Eiffel Tower?" The answer is ``Celebrating the 100th anniversary of the victory of the French Revolution''. If the input question changes to ``Will the Eiffel Tower fall down in the future?'', the system will return ``By the Seine''. While the answer to the latter question is incorrect, it can be seen as an entertainment scenario. Users can ask any question and get an interesting response based on the input document. 

\begin{figure}[h]
\centering
\includegraphics[scale=0.3]{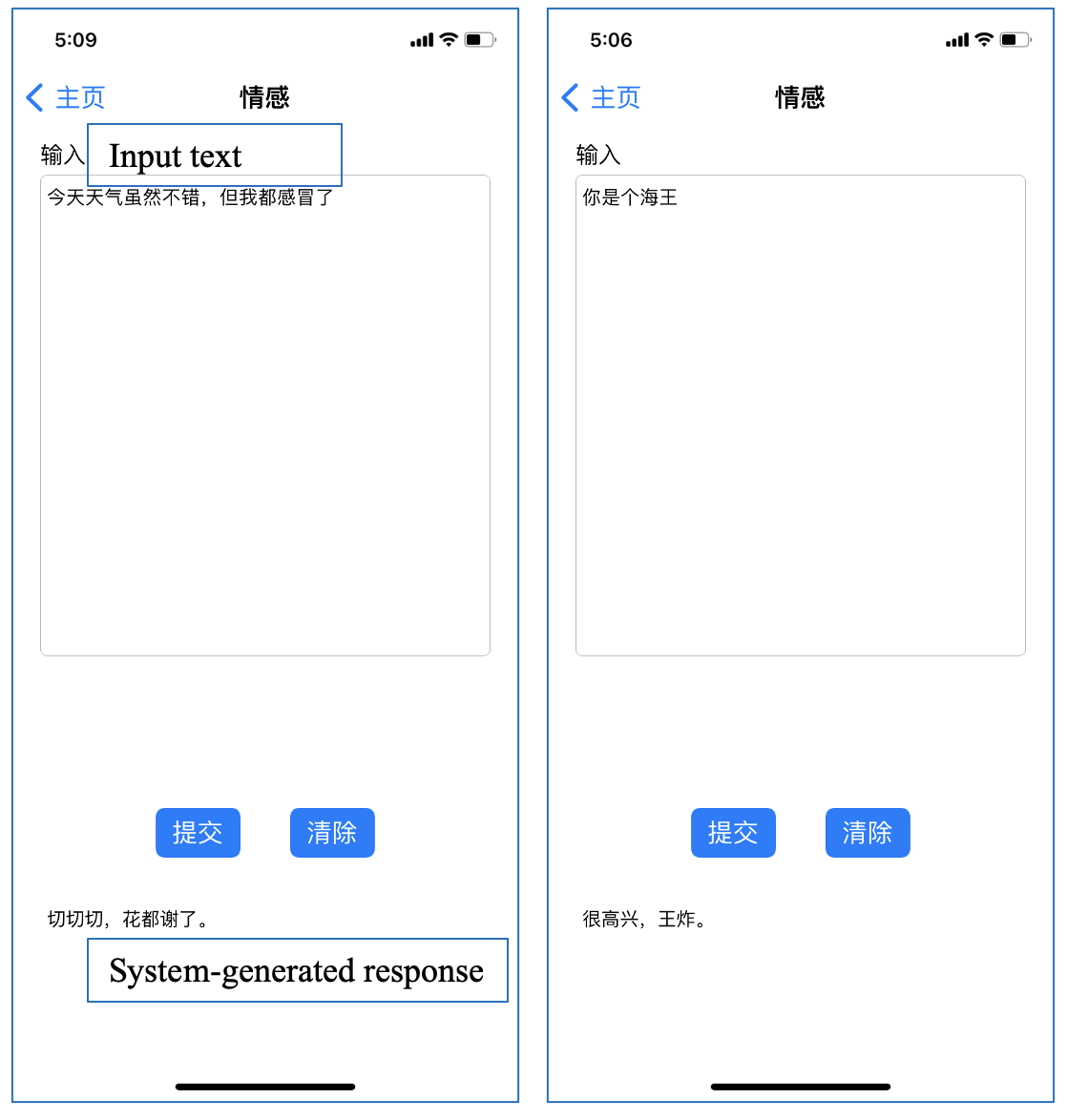}
\caption{Examples of emotion recognition function}
\label{fig:emotion}
\end{figure}

Figure \ref{fig:emotion} shows the results of emotion recognition. Emotions are divided into 5 levels, with the lowest level-1 representing the worst emotion and the highest level-5 representing the happiest emotion. When the user inputs ``The weather is fine today, but I have a cold.", the system responds with a sentiment score of 2. When the user inputs ``you are the sea king'', the system responds with a sentiment score of 5. Although the connotation of the latter sentence is derogatory, the system still judges it as a positive emotion. 

\subsection{Results of the Education Industry Application}
\begin{figure}[h]
\centering
\includegraphics[scale=0.4]{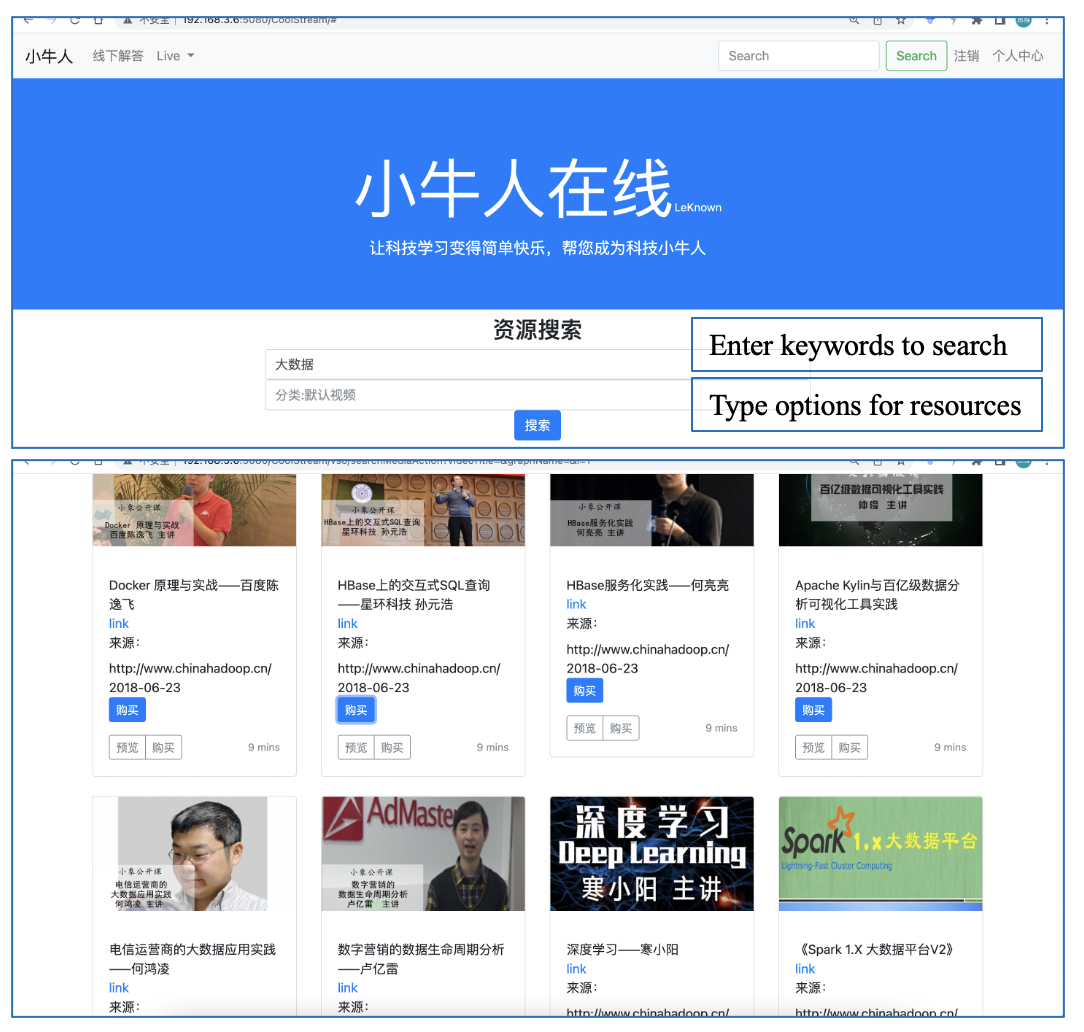}
\caption{The interface of our knowledge integration website}
\label{fig:xiaoniuren}
\end{figure}

For the knowledge integration website, we collect the educational resources of technologies such as big data, AI, machine learning, data mining, etc. on the Internet. We build a knowledge graph where users can search for resources through our websites. In the user center, users can upload new videos, download resources, play videos, manage resources, etc. As shown in Figure \ref{fig:xiaoniuren}, in the upper part of the diagram, the keywords and resource type entered are ``big data" and ``video" respectively, and then the site returns video tutorials related to big data. In the lower part of the diagram, the returned search results are displayed in a paginated manner, and users can bookmark these videos and watch them online.

\begin{figure}[h]
\centering
\includegraphics[scale=0.3]{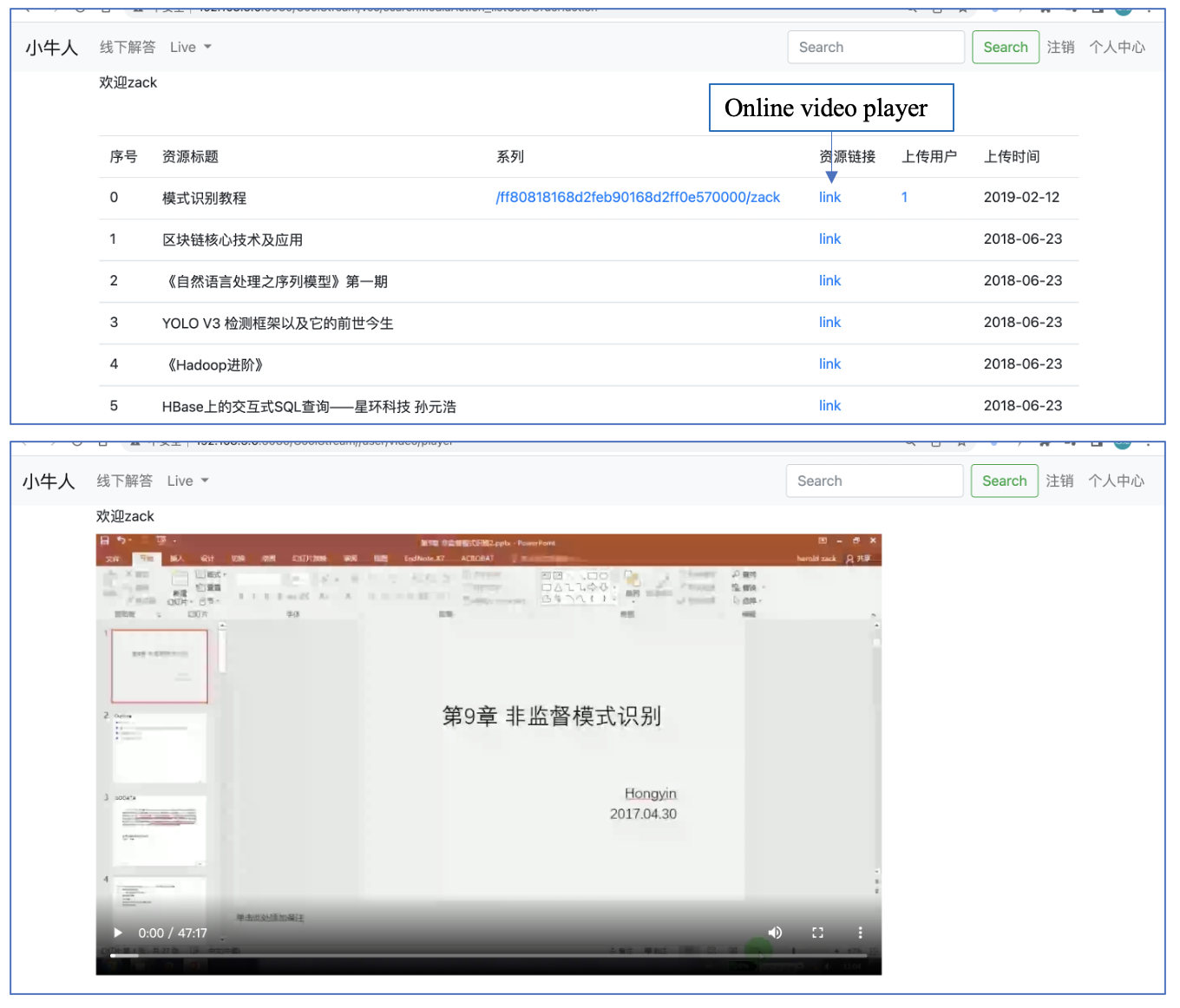}
\caption{The user center of our knowledge integration website}
\label{fig:xiaoniu2}
\end{figure}

In the user center, users can view their favorite videos and playlists, as shown in the upper part of Figure \ref{fig:xiaoniu2}. When the user clicks on a video playback link, videos uploaded by other users will be played on this site. For resources crawled on the Internet, users will be navigated to the target website after clicking, rather than playing on our website. As shown in the lower part of Figure \ref{fig:xiaoniu2}, the user is watching a video on pattern recognition - unsupervised pattern recognition.

\begin{figure}[h]
\centering
\includegraphics[scale=0.35]{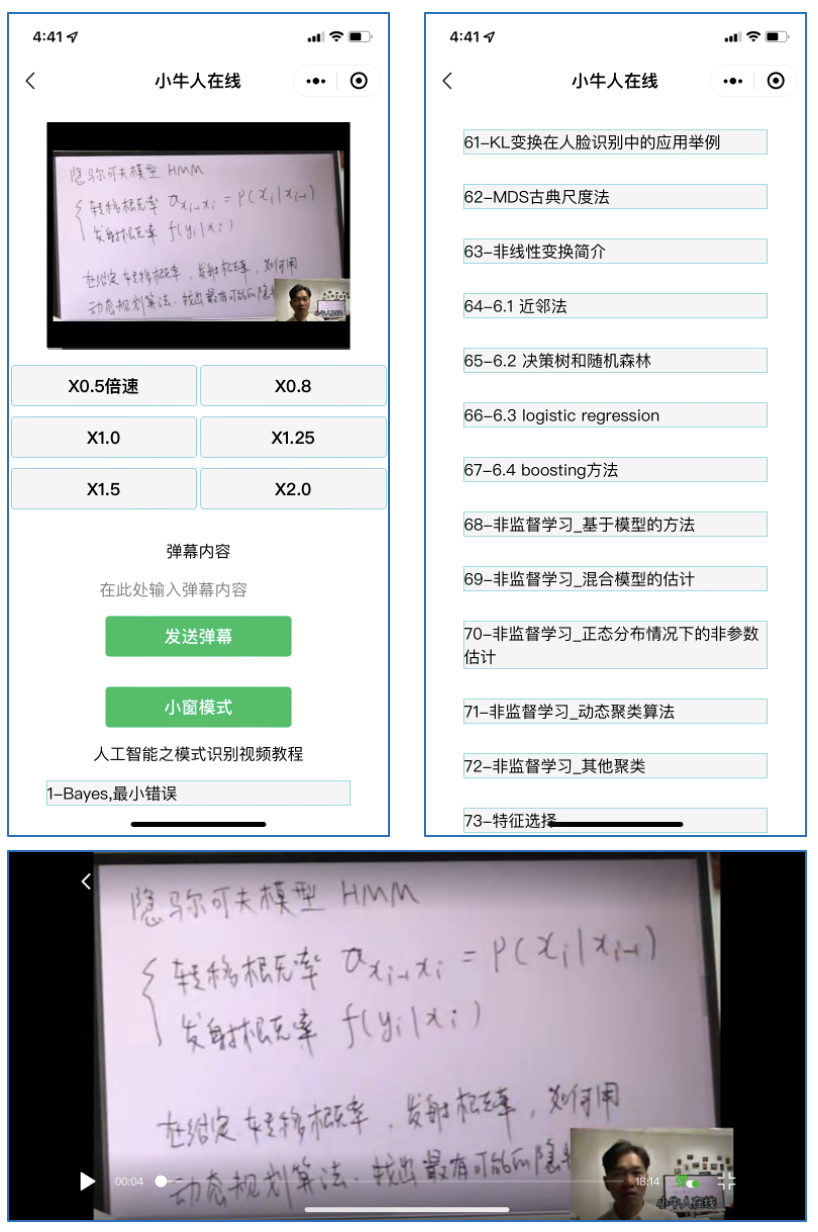}
\caption{The interface of our WeChat mini-program}
\label{fig:weichat}
\end{figure}

For the scarce resources on the Internet, we recorded video tutorials and wrote lectures. As shown in Figure \ref{fig:weichat}, the user chooses to watch the playlist of pattern recognition video tutorials through the WeChat mini-program. Users can watch playlists on several AI disciplines, such as pattern recognition, natural language processing, algorithm and data structure, matrix theory, probability theory and mathematical statistics, etc.

\begin{figure}[h]
\centering
\includegraphics[scale=0.3]{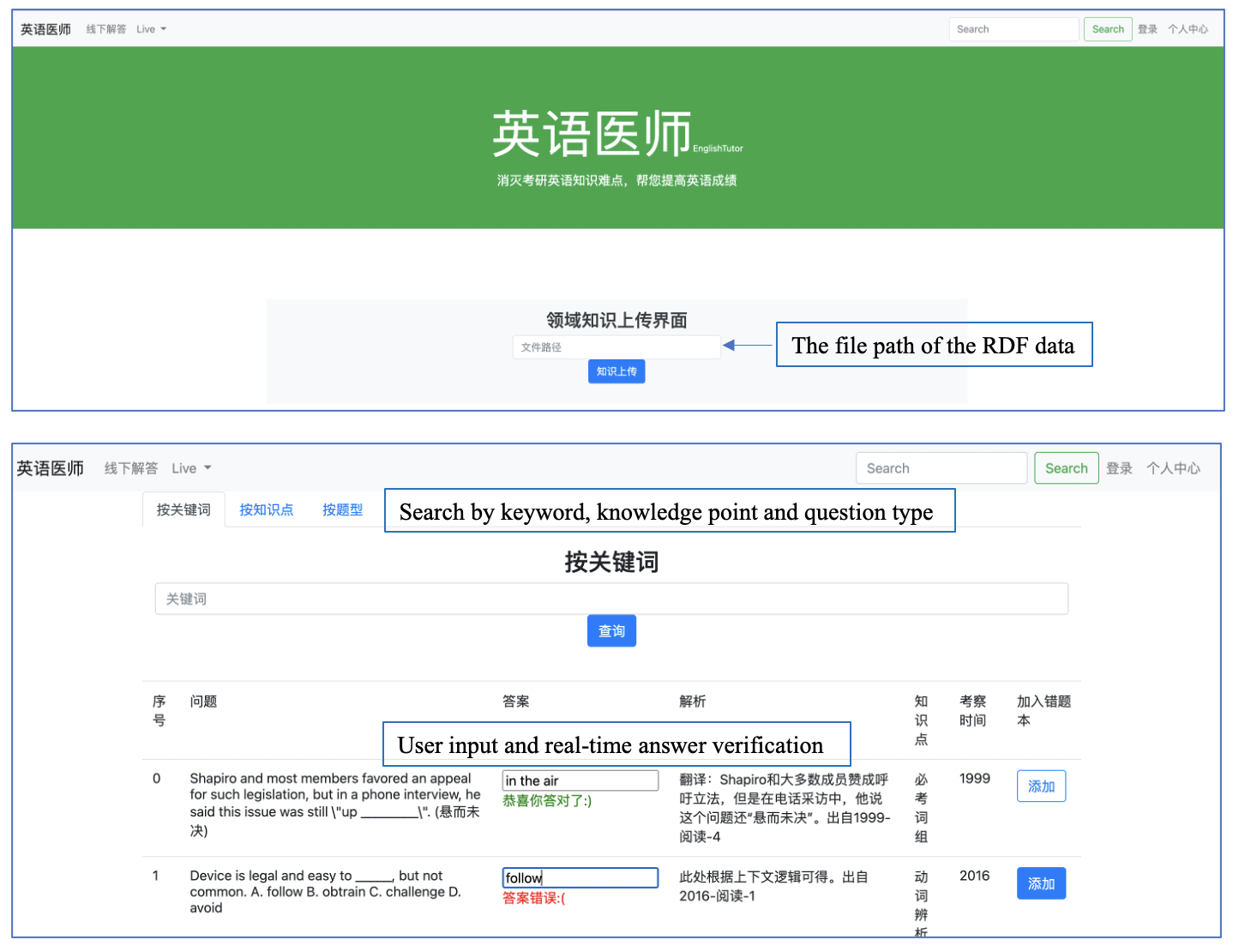}
\caption{The interface of our knowledge test website}
\label{fig:english}
\end{figure}

For knowledge testing, as shown in the upper part of Figure \ref{fig:english}, administrators can build their exam systems by generating RDF (Resource Description Framework in Semantic Web Standards) files and uploading them to the site. The lower part of the diagram shows that this website supports organizing questions by keyword, knowledge points, and question types. As the user fills in a slot, the system validates each answer in real-time and displays it below the question. 

\subsection{Results of the Consumption Industry Application}
\begin{figure}[h]
\centering
\includegraphics[scale=0.3]{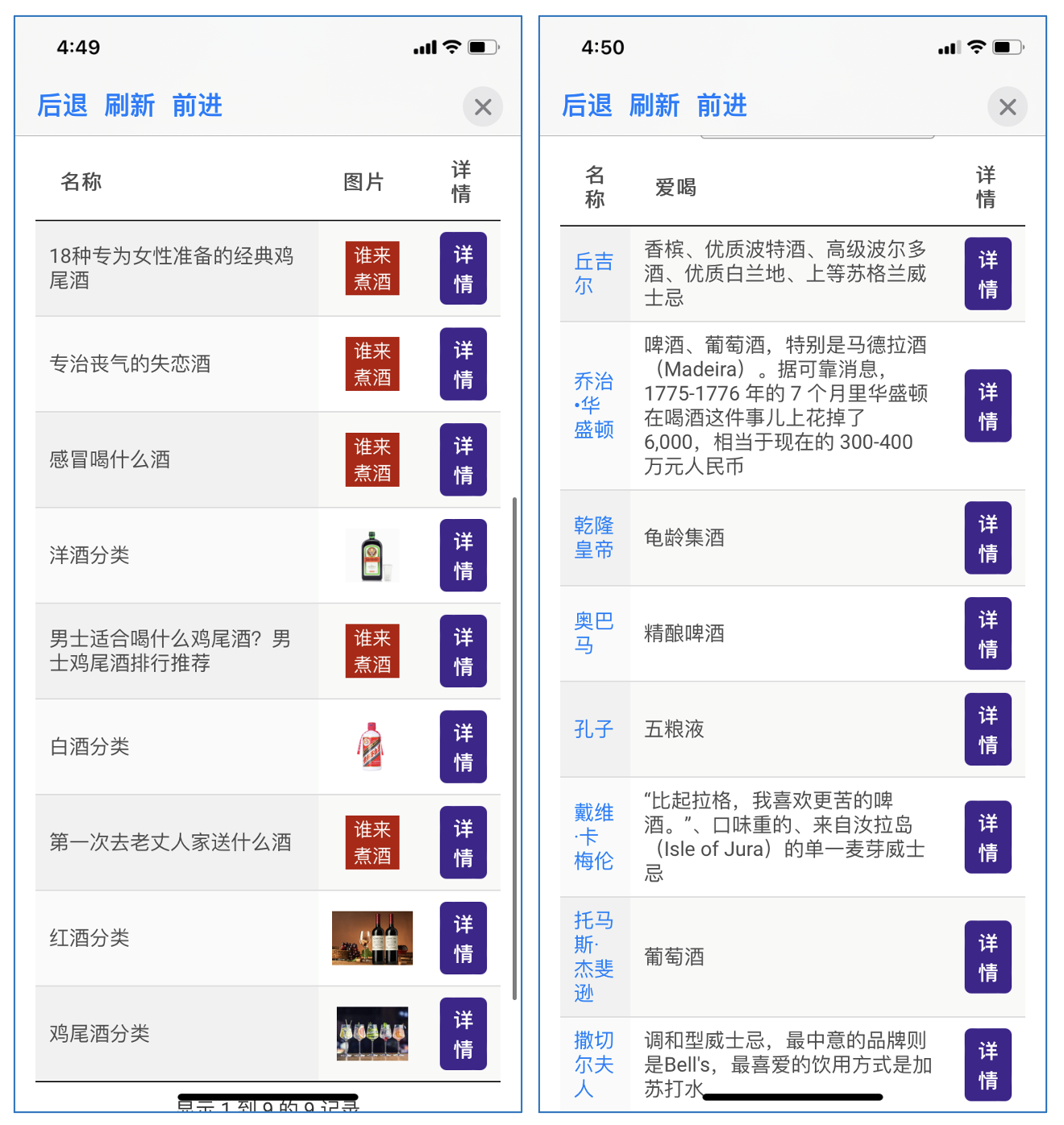}
\caption{The classification of alcoholic beverages (left) and celebrities and alcoholic drinks (right)}
\label{fig:consumption2}
\end{figure}

Figure \ref{fig:consumption2} shows the results of the ``Talk over drinks'' app. The left part of the diagram is the classification of alcoholic beverages, including liquor, red wine, imported wine, cocktails, etc. Users can click ``Details" to view more information and recommended reading content. The right part of the diagram is the alcoholic beverages that celebrities like to drink. Users can click on a celebrity to see information about that celebrity and click on ``Details" to see the story that the celebrity likes to drink. 

\begin{figure}[h]
\centering
\includegraphics[scale=0.35]{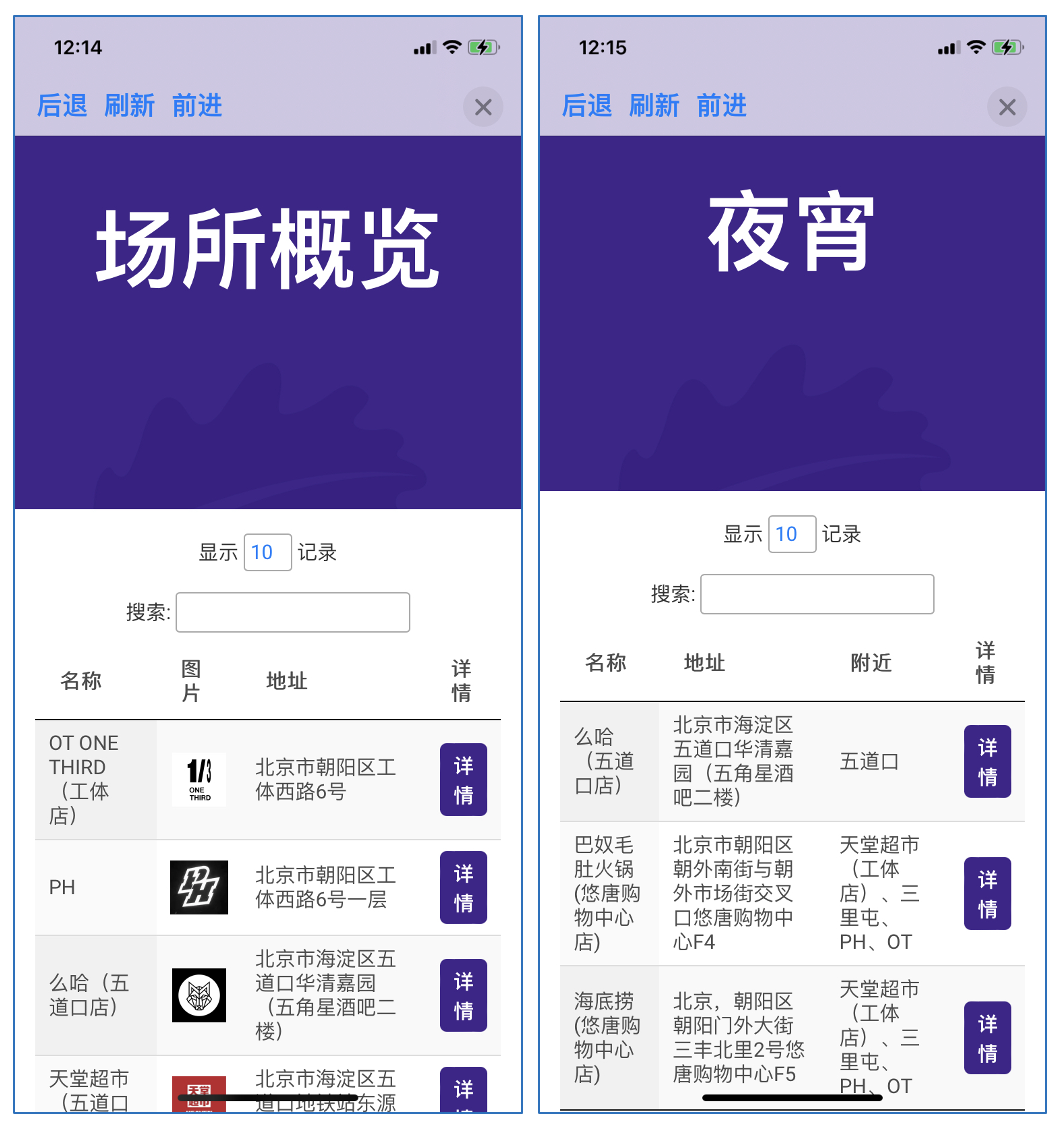}
\caption{The list of physical stores (left) and late night snacks (right)}
\label{fig:consumption1}
\end{figure}

As shown in Figure \ref{fig:consumption1}, the left part of the diagram shows an overview of the venue, i.e., pubs, and the right part of the diagram shows the restaurants serving late-night snacks. We reorganize the information about physical stores based on subway stations, bus stops, and consumption levels. Users can easily search for these physical stores by location, name, and consumption level, and can also view more information and reviews of these stores by clicking on the ``Details". There are buttons in the detailed descriptions to guide users to apps such as dianping.com, meituan.com, and WeChat official accounts.

\section{Conclusion and Future Work}
This paper proposes a flexible Metaverse AI technology framework MetaAID, which aims to support AI technology in the development of metaverse applications. This framework uses AI technology and human editing to support the collaborative development of applications in multiple industries. We apply the MetaAID framework to 3 industries that are indispensable for the expansion of domestic demand and economic internal circulation, i.e., entertainment (NLP-based entertainment), online education (AI subjects and English learning), daily consumption (food and beverages). The experimental results demonstrate that our framework can be used to support the development of metaverse applications.

The innovation of this paper lies in the flexibility of the framework and the diversity of techniques within this tech stack, including app development technologies, AI technologies, and human editing technologies, which can be used to build applications in different industries. Applications built on this framework can facilitate information exchange and federated data analysis. In the future, we hope to expand this framework to support scalable web architecture and distributed systems.

\end{CJK*}
\bibliographystyle{aaai}
\bibliography{reference}

\end{document}